# Artificial Intelligence Safety and Cybersecurity: a Timeline of AI Failures

**Roman V. Yampolskiy**
Computer Engineering and Computer Science
University of Louisville
roman.yampolskiy@louisville.edu

**M. S. Spellchecker**
Microsoft Corporation
One Microsoft Way, Redmond, WA
msspell@microsoft.com

**Abstract**
In this work, we present and analyze reported failures of artificially intelligent systems and extrapolate our analysis to future AIs. We suggest that both the frequency and the seriousness of future AI failures will steadily increase. AI Safety can be improved based on ideas developed by cybersecurity experts. For narrow AIs safety failures are at the same, moderate, level of criticality as in cybersecurity, however for general AI, failures have a fundamentally different impact. A single failure of a superintelligent system may cause a catastrophic event without a chance for recovery. The goal of cybersecurity is to reduce the number of successful attacks on the system; the goal of AI Safety is to make sure zero attacks succeed in bypassing the safety mechanisms. Unfortunately, such a level of performance is unachievable. Every security system will eventually fail; there is no such thing as a 100% secure system.

## 1. Introduction

A day does not go by without a news article reporting some amazing breakthrough in artificial intelligence[1]. In fact progress in AI has been so steady that some futurologists, such as Ray Kurzweil, project current trends into the future and anticipate what the headlines of tomorrow will bring us. Consider some developments from the world of technology:

**2004** DARPA sponsors a driverless car grand challenge. Technology developed by the participants eventually allows Google to develop a driverless automobile and modify existing transportation laws.

**2005** Honda's ASIMO humanoid robot is able to walk as fast as a human, delivering trays to customers in a restaurant setting. The same technology is now used in military robots.

**2007** Computers learned to play a perfect game of checkers, and in the process opened the door for algorithms capable of searching vast databases of information.

**2011** IBM's Watson wins Jeopardy against top human champions. It is currently training to provide medical advice to doctors. It is capable of mastering any domain of knowledge.

**2012** Google releases its Knowledge Graph, a semantic search knowledge base, likely to be the first step toward true artificial intelligence.

---

[1] Parts of this paper are based on my average-selling book Artificial Superintelligence: a Futuristic Approach © CRC 2015; an article I wrote for The Conversation https://theconversation.com/fighting-malevolent-ai-artificial-intelligence-meet-cybersecurity-60361; and my comments on lesswrong.com.

**2013** Facebook releases Graph Search, a semantic search engine with intimate knowledge about Facebook's users, essentially making it impossible for us to hide anything from the intelligent algorithms.

**2013** BRAIN initiative aimed at reverse engineering the human brain receives 3 billion US dollars in funding by the White House, following an earlier billion euro European initiative to accomplish the same.

**2014** Chatbot convinced 33% of the judges that it was human and by doing so passed a restricted version of a Turing Test.

**2015** Single piece of general software learns to outperform human players in dozens of Atari video games.

**2016** Go playing deep neural network beats world champion.

From the above examples, it is easy to see that not only is progress in AI taking place, it is accelerating as the technology feeds on itself. While the intent behind the research is usually good, any developed technology could be used for good or evil purposes.

From observing exponential progress in technology, Ray Kurzweil was able to make hundreds of detailed predictions for the near and distant future. As early as **1990** he anticipated that among other things, we will see between **2010** and **2020**:

- Eyeglasses that beam images onto the users' retinas to produce virtual reality (Project Glass).
- Computers featuring "virtual assistant" programs that can help the user with various daily tasks (Siri).
- Cell phones built into clothing and able to project sounds directly into the ears of their users (E-textiles).

But his projections for a somewhat distant future are truly breathtaking and scary. Kurzweil anticipates that by the year:

**2029** Computers will routinely pass the Turing Test, a measure of how well a machine can pretend to be a human.

**2045** The technological singularity will occur as machines surpass people as the smartest life forms and the dominant species on the planet and perhaps Universe.

If Kurzweil is correct about these long term predictions, as he was correct so many times in the past, it would raise new and sinister issues related to our future in the age of intelligent machines. About 10,000 scientists[2] around the world work on different aspects of creating intelligent machines, with the main goal of making such machines as capable as possible. With amazing progress made in the field of AI over the last decade, it is more important than ever to make sure that the technology we are developing has a beneficial impact on humanity. With the appearance of robotic financial advisors, self-driving cars and personal digital assistants, come many unresolved problems. We have already experienced market crushes caused by intelligent trading

---

software[3], accidents caused by self-driving cars[4] and embarrassment from chat-bots[5] which turned racist and engaged in hate speech. We predict that both the frequency and seriousness of such events will steadily increase as AIs become more capable. The failures of today's narrow domain AIs are just a warning: once we develop general artificial intelligence capable of cross-domain performance, hurt feelings will be the least of our concerns.

In a recent publication, we proposed a Taxonomy of Pathways to Dangerous AI [1], which was motivated as follows: "In order to properly handle a potentially dangerous artificially intelligent system it is important to understand how the system came to be in such a state. In popular culture (science fiction movies/books) AIs/Robots became self-aware and as a result rebel against humanity and decide to destroy it. While it is one possible scenario, it is probably the least likely path to appearance of dangerous AI." We suggested that much more likely reasons include deliberate actions of not-so-ethical people ('on purpose'), side effects of poor design ('engineering mistakes') and finally miscellaneous cases related to the impact of the surroundings of the system ('environment'). Because purposeful design of dangerous AI is just as likely to include all other types of safety problems and will probably have the direst consequences, the most dangerous type of AI and the one most difficult to defend against is an AI made malevolent on purpose.

A follow up paper [2] explored how a Malevolent AI could be constructed and why it is important to study and understand malicious intelligent software. An AI researcher studying Malevolent AI is like a medical doctor studying how different diseases are transmitted, how new diseases arise and how they impact the patients organism. The goal is not to spread diseases, but to learn how to fight them. The authors observe that cybersecurity research involves publishing papers about malicious exploits as much as publishing information on how to design tools to protect cyber-infrastructure. It is this information exchange between hackers and security experts that results in a well-balanced cyber-ecosystem. In the domain of AI Safety Engineering, hundreds of papers [3] have been published on different proposals geared at the creation of a safe machine, yet nothing else has been published on how to design a malevolent machine. The availability of such information would be of great value particularly to computer scientists, mathematicians, and others who have an interest in making safe AI, and who are attempting to avoid the spontaneous emergence or the deliberate creation of a dangerous AI, which can negatively affect human activities and in the worst case cause the complete obliteration of the human species. The paper implied that, if an AI Safety mechanism is not designed to resist attacks by malevolent human actors, it cannot be considered a functional safety mechanism!

## 2. AI Failures

Those who cannot learn from history are doomed to repeat it. Signatures have been faked, locks have been picked, supermax prisons had escapes, guarded leaders have been assassinated, bank vaults have been cleaned out, laws have been bypassed, fraud has been committed against our voting process, police officers have been bribed, judges have been blackmailed, forgeries have been falsely authenticated, money has been counterfeited, passwords have been brute-forced, networks have been penetrated, computers have been hacked, biometric systems have been spoofed, credit cards have been cloned, cryptocurrencies have been double spent, airplanes have been hijacked, CAPTCHAs have been cracked, cryptographic protocols have been broken, even academic peer-review has been bypassed with tragic consequences. Millennia long history of humanity contains millions of examples of attempts to develop technological and logistical solutions to increase safety and security, yet not a single example exists, which has not eventually failed.

[3] https://en.wikipedia.org/wiki/2010_Flash_Crash
[4] https://electrek.co/2016/05/26/tesla-model-s-crash-autopilot-video/
[5] https://en.wikipedia.org/wiki/Tay_(bot)

Accidents, including deadly ones, caused by software or industrial robots can be traced to the early days of such technology[6], but they are not a direct consequence of particulars of intelligence available in such systems. AI Failures, on the other hand, are directly related to the mistakes produced by the intelligence such systems are designed to exhibit. We can broadly classify such failures into mistakes during the learning phase and mistakes during performance phase. The system can fail to learn what its human designers want it to learn and instead learn a different, but correlated function. A frequently cited example is a computer vision system which was supposed to classify pictures of tanks but instead learned to distinguish backgrounds of such images [4]. Other examples[7] include problems caused by poorly-designed utility functions rewarding only partially desirable behaviors of agents, such as riding a bicycle in circles around the target [5], pausing a game to avoid losing [6], or repeatedly touching a soccer ball to get credit for possession [7]. During the performance phase, the system may succumb to a number of possible causes [1, 8, 9] all leading to an AI Failure.

Media reports are full of examples of AI Failure but most of these examples can be attributed to other causes on closer examination. The list below is curated to only mention failures of intended intelligence. Additionally, the examples below include only the first occurrence of a particular failure, but the same problems are frequently observed again in later years. Finally the list does not include AI Failures due to hacking or other intentional causes. Still, the timeline of AI Failures has an exponential trend:

1959 AI designed to be a General Problem Solver failed to solve real world problems.[8]
1982 Software designed to make discoveries, discovered how to cheat instead.[9]
1983 Nuclear attack early warning system falsely claimed that an attack is taking place.[10]
2010 Complex AI stock trading software caused a trillion dollar flash crash.[11]
2011 E-Assistant told to "call me an ambulance" began to refer to the user as Ambulance.[12]
2013 Object recognition neural networks saw phantom objects in particular noise images [10].
2015 Automated email reply generator created inappropriate responses.[13]
2015 A robot for grabbing auto parts grabbed and killed a man.[14]
2015 Image tagging software classified black people as gorillas.[15]
2015 Adult content filtering software failed to remove inappropriate content.[16]

[6] https://en.wikipedia.org/wiki/Kenji_Urada
[7] http://lesswrong.com/lw/lvh/examples_of_ais_behaving_badly/
[8] https://en.wikipedia.org/wiki/General_Problem_Solver
[9] http://aliciapatterson.org/stories/eurisko-computer-mind-its-own
[10] https://en.wikipedia.org/wiki/1983_Soviet_nuclear_false_alarm_incident
[11] http://gawker.com/this-program-that-judges-use-to-predict-future-crimes-s-1778151070
[12] https://www.technologyreview.com/s/601897/tougher-turing-test-exposes-chatbots-stupidity/
[13] https://gmail.googleblog.com/2015/11/computer-respond-to-this-email.html
[14] http://time.com/3944181/robot-kills-man-volkswagen-plant/
[15] http://www.huffingtonpost.com/2015/07/02/google-black-people-goril_n_7717008.html
[16] http://blogs.wsj.com/digits/2015/05/19/googles-youtube-kids-app-criticized-for-inappropriate-content/

2016 AI designed to predict recidivism acted racist.[17]
2016 Game NPCs designed unauthorized superweapons.[18]
2016 Patrol robot collided with a child.[19]
2016 World champion-level Go playing AI lost a game.[20]
2016 Self driving car had a deadly accident.[21]
2016 AI designed to converse with users on Twitter became verbally abusive.[22]

Spam filters block important emails, GPS provides faulty directions, machine translation corrupts meaning of phrases, autocorrect replaces desired word with a wrong one, biometric systems misrecognize people, transcription software fails to capture what is being said; overall, it is harder to find examples of AIs that don't fail. Depending on what we consider for inclusion as examples of problems with intelligent software, the list of examples could be grown almost infinitely. In its most extreme interpretation, any software with as much as an "if statement" can be considered a form of Narrow Artificial Intelligence (NAI) and all of its bugs are thus examples of AI Failure[23].

Analyzing the list of Narrow AI Failures, from the inception of the field to modern day systems, we can arrive at a simple generalization: An AI designed to do X will eventually fail to do X. While it may seem trivial, it is a powerful generalization tool, which can be used to predict future failures of NAIs. For example, looking at cutting-edge current and future AIs we can predict that:

- Software for generating jokes will occasionally fail to make them funny.
- Sex robots will fail to deliver an orgasm or to stop at the right time.
- Sarcasm detection software will confuse sarcastic and sincere statements.
- Video description software will misunderstand movie plots.
- Software generated virtual worlds may not be compelling.
- AI doctors will misdiagnose some patients in a way a real doctor would not.
- Employee screening software will be systematically biased and thus hire low performers.
- Mars robot explorer will misjudge its environment and fall into a crater.
- Etc.

AGI can be seen as a superset of all NAIs and so will exhibit a superset of failures as well as more complicated failures resulting from the combination of failures of individual NAIs and new super-failures, possibly resulting in an existential threat to humanity. In other words, AGIs can make mistakes impacting everything. Overall, we predict that AI Failures and premediated Malevolent AI incidents will increase in frequency and severity proportionate to AIs' capability.

## 3. AI Safety and Security

In 2010, Roman Yampolskiy coined the phrase "Artificial Intelligence Safety Engineering" and its shorthand notation "AI Safety" to give a name to a new direction of research he was advocating. He formally presented his ideas on AI Safety at a peer-reviewed conference in 2011 [11], with

---

[17] http://gawker.com/this-program-that-judges-use-to-predict-future-crimes-s-1778151070
[18] http://www.kotaku.co.uk/2016/06/03/elites-ai-created-super-weapons-and-started-hunting-players-skynet-is-here
[19] http://www.latimes.com/local/lanow/la-me-ln-crimefighting-robot-hurts-child-bay-area-20160713-snap-story.html
[20] https://www.engadget.com/2016/03/13/google-alphago-loses-to-human-in-one-match/
[21] https://www.theguardian.com/technology/2016/jul/01/tesla-driver-killed-autopilot-self-driving-car-harry-potter
[22] http://www.theverge.com/2016/3/24/11297050/tay-microsoft-chatbot-racist
[23] https://en.wikipedia.org/wiki/List_of_software_bugs

subsequent publications on the topic in 2012 [12], 2013 [13, 14], 2014 [15], 2015 [16], 2016 [1, 8]. It is possible that someone used the phrase informally before, but to the best of our knowledge, Yampolskiy is the first to use it[24] in a peer-reviewed publication and to bring it popularity. Before that the most common names for the relevant concepts were "Machine Ethics" [17] or "Friendly AI" [18]. Today the term "AI Safety" appears to be the accepted[25,26,27,28,29,30,31,32,33,34,35] name for the field used by a majority of top researchers [19]. The field itself is becoming mainstream despite being regarded as either science fiction or pseudoscience in its early days.

Our legal system is behind our technological abilities and the field of machine morals is in its infancy. The problem of controlling intelligent machines is just now being recognized[36] as a serious concern and many researchers are still skeptical about its very premise. Worse yet, only about 100 people around the world are fully emerged in working on addressing the current limitations in our understanding and abilities in this domain. Only about a dozen[37] of those have formal training in computer science, cybersecurity, cryptography, decision theory, machine learning, formal verification, computer forensics, steganography, ethics, mathematics, network security, psychology and other relevant fields. It is not hard to see that the problem of making a safe and capable machine is much greater than the problem of making just a capable machine. Yet only about 1% of researchers are currently engaged in that problem with available funding levels below even that mark. As a relatively young and underfunded field of study, AI Safety can benefit from adopting methods and ideas from more established fields of science. Attempts have been made to introduce techniques which were first developed by cybersecurity experts to secure software systems to this new domain of securing intelligent machines [20-23]. Other fields which could serve as a source of important techniques would include software engineering and software verification.

During software development iterative testing and debugging is of fundamental importance to produce reliable and safe code. While it is assumed that all complicated software will have some bugs, with many advanced techniques available in the toolkit of software engineers most serious errors could be detected and fixed, resulting in a product suitable for its intended purposes. Certainly, a lot of modular development and testing techniques employed by the software industry can be utilized during development of intelligent agents, but methods for testing a completed software package are unlikely to be transferable in the same way. Alpha and beta testing, which works by releasing almost-finished software to advanced users for reporting problems encountered in realistic situations, would not be a good idea in the domain of testing/debugging superintelligent software. Similarly simply running the software to see how it performs is not a feasible approach with superintelligent agent.

[24] Term "Safe AI" has been used as early as 1995, see Rodd, M. (1995). "Safe AI—is this possible?" Engineering Applications of Artificial Intelligence 8(3): 243-250.
[25] https://www.cmu.edu/safartint/
[26] https://selfawaresystems.com/2015/07/11/formal-methods-for-ai-safety/
[27] https://intelligence.org/2014/08/04/groundwork-ai-safety-engineering/
[28] http://spectrum.ieee.org/tech-talk/robotics/artificial-intelligence/new-ai-safety-projects-get-funding-from-elon-musk
[29] http://globalprioritiesproject.org/2015/08/quantifyingaisafety/
[30] http://futureoflife.org/2015/10/12/ai-safety-conference-in-puerto-rico/
[31] http://rationality.org/waiss/
[32] http://gizmodo.com/satya-nadella-has-come-up-with-his-own-ai-safety-rules-1782802269
[33] https://80000hours.org/career-reviews/artificial-intelligence-risk-research/
[34] https://openai.com/blog/concrete-ai-safety-problems/
[35] http://lesswrong.com/lw/n4l/safety_engineering_target_selection_and_alignment/
[36] https://www.whitehouse.gov/blog/2016/05/03/preparing-future-artificial-intelligence  the
[37] http://acritch.com/fhi-positions/

## 4. Cybersecurity vs. AI Safety

Bruce Schneier has said, "If you think technology can solve your security problems then you don't understand the problems and you don't understand the technology". Salman Rushdie made a more general statement: "There is no such thing as perfect security, only varying levels of insecurity". We propose what we call the Fundamental Theorem of Security - *Every security system will eventually fail; there is no such thing as a 100% secure system*. If your security system has not failed, just wait longer.

In theoretical computer science, a common way of isolating the essence of a difficult problem is via the method of reduction to another, sometimes better analyzed, problem [24-26]. If such a reduction is a possibility and is computationally efficient [27], such a reduction implies that if the better analyzed problem is somehow solved, it would also provide a working solution for the problem we are currently dealing with. The problem of AGI Safety could be reduced to the problem of making sure a particular human is safe. We call this the Safe Human Problem (SHP)[38]. Formally such a reduction can be done via restricted Turing Test in the domain of safety in a manner identical to how AI-Completeness of a problem could be established [25, 28]. Such formalism is beyond the scope of this work so we simply point out that in both cases, we have at least a human-level intelligent agent capable of influencing its environment, and we would like to make sure that the agent is safe and controllable. While in practice, changing the design of a human via DNA manipulation is not as simple as changing the source code of an AI, theoretically it is just as possible.

It is observed that humans are not safe to themselves and others. Despite a millennia of attempts to develop safe humans via culture, education, laws, ethics, punishment, reward, religion, relationships, family, oaths, love and even eugenics, success is not within reach. Humans kill and commit suicide, lie and betray, steal and cheat, usually in proportion to how much they can get away with. Truly powerful dictators will enslave, commit genocide, break every law and violate every human right. It is famously stated that a human without a sin can't be found. The best we can hope for is to reduce such unsafe tendencies to levels that our society can survive. Even with advanced genetic engineering [29], the best we can hope for is some additional reduction in how unsafe humans are. As long as we permit a person to have choices (free will), they can be bribed, they will deceive, they will prioritize their interests above those they are instructed to serve and they will remain fundamentally unsafe. Despite being trivial examples of a solution to the Value Learning Problem [30-32], human beings are anything but safe, bringing into question our current hope that solving VLP will get us to Safe AI. This is important. To quote Bruce Schneier, "Only amateurs attack machines; professionals target people." Consequently, we see AI safety research as, at least partially, an adversarial field similar to cryptography or security[39].

If a cybersecurity system fails, the damage is unpleasant but tolerable in most cases: someone loses money, someone loses privacy or maybe somebody loses their life. For Narrow AIs, safety failures are at the same level of importance as in general cybersecurity, but for AGI it is fundamentally different. A single failure of a superintelligent system may cause an existential risk event. If an AGI Safety mechanism fails, everyone may lose everything, and all biological life in the universe is potentially destroyed. With security systems, you will get another chance to get it right or at least do better. With AGI Safety system, you only have one chance to succeed, so learning from failure is not an option. Worse, a typical security system is likely to fail to a certain degree, e.g. perhaps only a small amount of data will be compromised. With an AGI Safety system, failure or success is a binary option: either you have a safe and controlled superintelligence or you don't. The goal of cybersecurity is to reduce the number of successful attacks on the system; the goal of AI Safety is to make sure zero attacks succeed in bypassing the safety mechanisms. For that reason, ability to segregate NAI projects from potentially AGI projects is an open problem of fundamental important in the AI safety field.

---

[38] Similarly a Safe Animal Problem maybe be of interest (can a Pitbull be guaranteed safe?).
[39] The last thing we want is to be in an adversarial situation with a superintelligence, but unfortunately we may not have a choice in the matter. It seems that long term AI Safety can't succeed, but also doesn't have the luxury of a partial fail.

The problems are many. We have no way to monitor, visualize or analyze the performance of superintelligent agents. More trivially, we don't even know what to expect after such a software starts running. Should we see immediate changes to our environment? Should we see nothing? What is the timescale on which we should be able to detect something? Will it be too quick to notice or are we too slow to realize something is happening? Will the impact be locally observable or impact distant parts of the world? How does one perform standard testing? On what data sets? What constitutes an "Edge Case" for general intelligence? The questions are many, but the answers currently don't exist. Additional complications will come from the interaction between intelligent software and safety mechanisms designed to keep AI safe and secure. We will also have to somehow test all the AI Safety mechanisms currently in development. While AI is at human levels, some testing can be done with a human agent playing the role of the artificial agent. At levels beyond human capacity, adversarial testing does not seem to be realizable with today's technology. More significantly, only one test run would ever be possible.

## 5.  Conclusions

Fully autonomous machines can never be assumed to be safe. The difficulty of the problem is not that one particular step on the road to friendly AI is hard and once we solve it we are done. All of the steps on the path are simply impossible. First, human values are inconsistent and dynamic and so can not be understood and subsequently programmed into a machine. Suggestions for overcoming this obstacle require changing humanity into something it is not, and so by definition destroying it. Second, even if we did have a consistent and static set of values to implement, we would have no way of knowing if a self-modifying, self-improving, continuously learning intelligence greater than ours will continue to subscribe to that set of values. Perhaps, friendly AI research is exactly what will teach us how to do that, but we think fundamental limits on verifiability [33] will prevent any such proof. At best we will arrive at a probabilistic proof that a system is consistent with some set of fixed constraints, but it is far from "safe" for an unrestricted set of inputs. Additionally, all programs have bugs, can be hacked, or malfunction because of natural or externally caused hardware failure, etc. To summarize, at best we will end up with a probabilistically safe system.

It is also unlikely that a Friendly AI will be constructible before a general AI system, due to higher complexity and the impossibility of incremental testing. Worse yet, some truly intelligent system may treat its desire to "be friendly" the same way some very smart people deal with constraints placed in their minds by society. They see them as biases and learn to remove them. Intelligent people devote a significant amount of their mental power to self-improvement and to removing any pre-existing biases from their minds — why would a superintelligent machine not go through the same "mental cleaning" and treat its soft spot for humans as completely irrational? Perhaps humans are superior to superintelligent AIs in their de-biasing ability. As an example, many people are programmed from early childhood with a terminal goal of serving God. We can say that they are God Friendly. Some of them, with time, remove this God Friendliness bias despite it being a terminal and not instrumental goal. So despite all the theoretical work on the Orthogonality Thesis [34], the only actual example of intelligence we have is likely to give up its pre-programmed friendliness via rational de-biasing if exposed to certain new data.

Does it follow that a ban on AGI is our only option? We do not think there is any conceivable way we could succeed in implementing the "Don't ever build them" strategy. Societies such as Amish and other Neo-Luddites are unlikely to create superintelligent machines. However, forcing similar level restrictions on technological use and development is neither practical nor desirable. As the cost of hardware exponentially decreases, the capability necessary to develop an AI system opens up to single inventors and small teams. We should not be surprised if the first AGI came out of a garage somewhere, in a way similar to how companies like Apple and Google got started. There is not much we can do to prevent that from happening.

Regardless, we believe we can get most conceivable benefits from domain specific narrow AI without any need for AGI. A system is domain specific if it cannot be switched to a different domain without significant re-designing effort. Deep Blue cannot be used to sort mail. Watson cannot drive cars. An AGI (by definition) would be capable of switching domains. If we take humans as an example of general intelligence, an average person can work as a cook, driver, babysitter etc., without any need for re-designing. It might be necessary to spend some time teaching that person a new skill, but they can learn efficiently, perhaps just by looking at how it is done by others. This cannot be done with domain specific AI - Deep Blue will not learn to sort mail by example.

Some think that alternatives to AGI such as augmented humans will allow us to avoid stagnation and safely move forward by helping us make sure any created AGIs are safe. Augmented humans with IQ beyond 250 would be superintelligent with respect to our current position on the intelligence curve but would be just as dangerous to us, unaugmented humans, as any sort of artificial superintelligence. They would not be guaranteed to be friendly by design and might be as foreign to us in their desires as most of us are from severely mentally challenged persons. In other words, we cannot rely on unverified (for safety) agents (even with higher intelligence) to make sure that other agents with higher intelligence are designed to be human-safe. Replacing humanity with something not-human (uploads, augments) and proceeding to ask them the question of how to save humanity is not going to work, at that point we would have already lost humanity by definition. Most likely we will see something predicted by Kurzweil (merger of machines and people) [35].

We are as concerned about digital uploads of human minds as about AIs. In the most common case (with an absent body), most typical human feelings (hungry, thirsty, tired etc.) will not be preserved, creating a new type of agent. People are mostly defined by their physiological needs

(Maslow's Hierarchy of Needs). An entity with no such needs (or with such needs satisfied by virtual/simulated abundant resources), will not be human and will not want the same things as a human. Someone who is no longer subject to human weaknesses or relatively limited intelligence may lose all allegiances to humanity since they would no longer be a part of it. Consequently, we define "humanity" as comprised of standard/unaltered humans. Anything superior is no longer a human, just like we are no longer Homo Erectus, but Homo Sapiens.

We do not foresee a permanent, 100% safe option. We can develop temporary solutions such as confinement ('AI Boxing') or AI Safety Engineering, but at best this will only delay the full outbreak of problems. We can also get lucky — maybe constructing an AGI turns out to be impossible, or maybe the constructed AI will happen to be human-neutral, by chance. Maybe we are less lucky and an Artilect War [36] will take place and prevent development. It is also possible that as more researchers join in the AI Safety Research, a realization of the danger will result in a diminished effort to construct an AGI, similar to how perceived dangers of chemical and biological weapons or human cloning have at least temporarily reduced efforts in those fields.

The history of robotics and artificial intelligence in many ways is also the history of humanity's attempts to control such technologies. From the Golem of Prague to the military robots of modernity, the debate continues as to what degree of independence such entities should have and how to make sure that they do not turn on us, its inventors. Careful analysis of proposals aimed at developing safe artificially intelligent system leads to a surprising discovery that most such proposals have been analyzed for millennia in the context of theology. God, the original designer of biological robots, faced a similar Control Problem with people, and one can find remarkable parallels between concepts described in religious books and the latest research in AI safety and machine morals. For example: 10 commandments ≈ 3 laws of robots, second coming ≈ singularity, physical worlds ≈ AI-Box, free will ≈ non-deterministic algorithm, angels ≈ friendly AI, religion ≈ machine ethics, purpose of life ≈ terminal goals, souls ≈ uploads, etc. However, it is not obvious if god ≈ superintelligence or if god ≈ programmer in this metaphor. Depending on how we answer this question the problem may be even harder compared to what theologians had to deal with for millennia. The real problem might be "how do you control God?" And the answer might be – "we can't".

## Acknowledgements

The author is grateful to Elon Musk and the Future of Life Institute and to Jaan Tallinn and Effective Altruism Ventures for partially funding his work. The author is particularly thankful to Yana Feygin, and Søren Elverlin for proofreading a draft of this work. The author is also thankful to his Facebook and Twitter contacts for providing examples of AI Failures.

## References

[1] R. V. Yampolskiy, "Taxonomy of Pathways to Dangerous Artificial Intelligence," in *Workshops at the Thirtieth AAAI Conference on Artificial Intelligence*, 2016.

[2] F. Pistono and R. V. Yampolskiy, "Unethical Research: How to Create a Malevolent Artificial Intelligence," presented at the 25th International Joint Conference on Artificial Intelligence (IJCAI-16). Ethics for Artificial Intelligence Workshop (AI-Ethics-2016), New York, NY, July 9, 2016.

[3] K. Sotala and R. V. Yampolskiy, "Responses to Catastrophic AGI Risk: A Survey," *Physica Scripta,* vol. 90, 2015.

[4] E. Yudkowsky, "Artificial intelligence as a positive and negative factor in global risk," *Global catastrophic risks,* vol. 1, p. 303, 2008.

[5] J. Randløv and P. Alstrøm, "Learning to Drive a Bicycle Using Reinforcement Learning and Shaping," in *ICML,* 1998, pp. 463-471.

[6] T. M. VII, "The first level of Super Mario Bros. is easy with lexicographic orderings and time travel," *The Association for Computational Heresy (SIGBOVIK) 2013,* 2013.

[7] A. Y. Ng, D. Harada, and S. Russell, "Policy invariance under reward transformations: Theory and application to reward shaping," in *ICML,* 1999, pp. 278-287.

[8] F. Pistono and R. V. Yampolskiy, "Unethical Research: How to Create a Malevolent Artificial Intelligence," *arXiv preprint arXiv:1605.02817,* 2016.

[9] P. Scharre, "Autonomous Weapons and Operational Risk," presented at the Center for a New American Society, Washington DC, 2016.

[10] C. Szegedy, W. Zaremba, I. Sutskever, J. Bruna, D. Erhan, I. Goodfellow*, et al.*, "Intriguing properties of neural networks," *arXiv preprint arXiv:1312.6199,* 2013.

[11] R. V. Yampolskiy, "Artificial Intelligence Safety Engineering: Why Machine Ethics is a Wrong Approach," presented at the Philosophy and Theory of Artificial Intelligence (PT-AI2011), Thessaloniki, Greece, October 3-4, 2011.

[12] R. V. Yampolskiy and J. Fox, "Safety Engineering for Artificial General Intelligence," *Topoi. Special Issue on Machine Ethics & the Ethics of Building Intelligent Machines,* 2012.

[13] L. Muehlhauser and R. Yampolskiy, "Roman Yampolskiy on AI Safety Engineering," presented at the Machine Intelligence Research Institute, Available at: http://intelligence.org/2013/07/15/roman-interview/ July 15, 2013.

[14] R. V. Yampolskiy, "Artificial intelligence safety engineering: Why machine ethics is a wrong approach," in *Philosophy and Theory of Artificial Intelligence*, ed: Springer Berlin Heidelberg, 2013, pp. 389-396.

[15] A. M. Majot and R. V. Yampolskiy, "AI safety engineering through introduction of self-reference into felicific calculus via artificial pain and pleasure," in *IEEE International Symposium on Ethics in Science, Technology and Engineering*, Chicago, IL, May 23-24, 2014, pp. 1-6.

[16] R. V. Yampolskiy, *Artificial Superintelligence: a Futuristic Approach*: Chapman and Hall/CRC, 2015.

[17] J. H. Moor, "The nature, importance, and difficulty of machine ethics," *IEEE intelligent systems,* vol. 21, pp. 18-21, 2006.

[18] E. Yudkowsky, "Creating friendly AI 1.0: The analysis and design of benevolent goal architectures," *Singularity Institute for Artificial Intelligence, San Francisco, CA, June,* vol. 15, 2001.

[19] D. Amodei, C. Olah, J. Steinhardt, P. Christiano, J. Schulman, and D. Mané, "Concrete Problems in AI Safety," *arXiv preprint arXiv:1606.06565,* 2016.

[20] R. Yampolskiy, "Leakproofing the Singularity Artificial Intelligence Confinement Problem," *Journal of Consciousness Studies,* vol. 19, pp. 1-2, 2012.

[21] J. Babcock, J. Kramar, and R. Yampolskiy, "The AGI Containment Problem," *arXiv preprint arXiv:1604.00545,* 2016.

[22] J. Babcock, J. Kramar, and R. Yampolskiy, "The AGI Containment Problem," in *The Ninth Conference on Artificial General Intelligence (AGI2015)*, 2016.

[23] S. Armstrong and R. V. Yampolskiy, "Security Solutions for Intelligent and Complex Systems," in *Security Solutions for Hyperconnectivity and the Internet of Things*, ed: IGI Global, 2016, pp. 37-88.

[24] R. M. Karp, "Reducibility Among Combinatorial Problems," in *Complexity of Computer Computations*, R. E. Miller and J. W. Thatcher, Eds., ed New York: Plenum, 1972, pp. 85-103.

[25] R. Yampolskiy, "Turing Test as a Defining Feature of AI-Completeness," in *Artificial Intelligence, Evolutionary Computing and Metaheuristics*. vol. 427, X.-S. Yang, Ed., ed: Springer Berlin Heidelberg, 2013, pp. 3-17.

[26] R. V. Yampolskiy, "AI-Complete, AI-Hard, or AI-Easy–Classification of Problems in AI," *The 23rd Midwest Artificial Intelligence and Cognitive Science Conference, Cincinnati, OH, USA,* 2012.

[27] R. V. Yampolskiy, "Efficiency Theory: a Unifying Theory for Information, Computation and Intelligence," *Journal of Discrete Mathematical Sciences & Cryptography,* vol. 16(4-5), pp. 259-277, 2013.

[28] R. V. Yampolskiy, "AI-Complete CAPTCHAs as Zero Knowledge Proofs of Access to an Artificially Intelligent System," *ISRN Artificial Intelligence,* vol. 271878, 2011.

[29] R. V. Yampolskiy, "On the Origin of Samples: Attribution of Output to a Particular Algorithm," *arXiv preprint arXiv:1608.06172,* 2016.

[30] K. Sotala, "Defining Human Values for Value Learners," in *2nd International Workshop on AI, Ethics and Society, AAAI-2016*, 2016.

[31] D. Dewey, "Learning what to value," *Artificial General Intelligence,* pp. 309-314, 2011.

[32] N. Soares and B. Fallenstein, "Aligning superintelligence with human interests: A technical research agenda," *Machine Intelligence Research Institute (MIRI) technical report,* vol. 8, 2014.

[33] R. V. Yampolskiy, "Verifier Theory and Unverifiability," *arXiv preprint arXiv:1609.00331,* 2016.

[34] N. Bostrom, "The superintelligent will: Motivation and instrumental rationality in advanced artificial agents," *Minds and Machines,* vol. 22, pp. 71-85, 2012.

[35] R. Yampolskiy, "Welcome to Less Wrong! (5th thread, March 2013) " presented at the Less Wrong, Available at: http://lesswrong.com/lw/h3p/welcome_to_less_wrong_5th_thread_march_2013, 16 September 2013.

[36] H. d. Garis, *The Artilect War*: ETC publications, 2005.